\documentclass[final]{cvpr}

\usepackage{times}
\usepackage{epsfig}
\usepackage{graphicx}
\usepackage{amsmath}
\usepackage{amssymb}
\usepackage{booktabs}
\usepackage{caption}
\usepackage{dirtytalk}
\usepackage{subcaption}
\usepackage{balance}

\usepackage{hyperref}
\hypersetup{colorlinks}

\def\cvprPaperID{46} 
\def\confYear{CVPR 2021}

\begin{document}

\title{Photozilla: A Large-Scale Photography Dataset and Visual Embedding for 20 Photography Styles}

\author{Trisha Singhal$^{1}$\\
{\tt\small trisha\_singhal@sutd.edu.sg}
\and
Junhua Liu$^{1,2}$\\
{\tt\small junhua\_liu@mymail.sutd.edu.sg, j@forth.ai}

\and
Lucienne T.M. Blessing$^{1}$\\
{\tt\small lucienne\_blessing@sutd.edu.sg}

\and
Kwan Hui Lim$^{1}$\\
{\tt\small kwanhui\_lim@sutd.edu.sg}\\
\and
$^{1}${Singapore University of Technology and Design, Singapore} \\
$^{2}${Forth AI, Singapore}}

\maketitle

\begin{abstract}
   The advent of social media platforms has been a catalyst for the development of digital photography that engendered a boom in vision applications. With this motivation, we introduce a large-scale dataset termed `Photozilla', which includes over $990k$ images belonging to $10$ different photographic styles. The dataset is then used to train $3$ classification models to automatically classify the images into the relevant style which resulted in an accuracy of $\sim96\%$. With the rapid evolution of digital photography, we have seen new types of photography styles emerging at an exponential rate. On that account, we present a novel Siamese-based network that uses the trained classification models as the base architecture to adapt and classify unseen styles with only $25$ training samples. We report an accuracy of over $68\%$ for identifying $10$ other distinct types of photography styles. \textbf{This dataset can be found at \url{https://trisha025.github.io/Photozilla/}}
\end{abstract}

\section{Introduction}
\begin{figure*}[h]
    \centering
    \includegraphics[width=\linewidth]{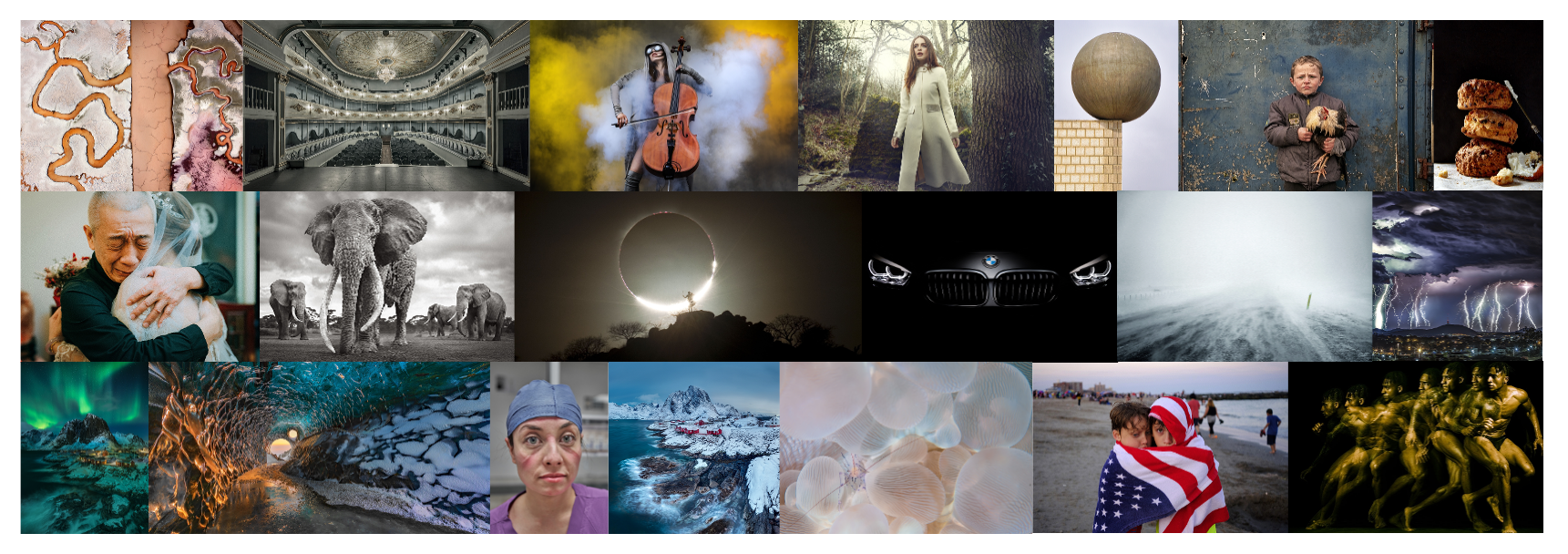}
    \caption{Awarded photos in different photographic styles. \\ \textit{(Source: International Photography Awards, 2020)}}
    \label{fig:top20}
\end{figure*}

\say{\textit{A picture is worth a thousand words.}}\textit{ - Henrik Ibsen}

\vspace{0.4cm}

Photography has become an integral part of people's everyday life for both professional and recreational purposes. Professionally, businesses use photographs for marketing and advertising purposes and for sharing news about significant political events. For recreational purposes, people use photographs to signify a purpose, such as telling a story, recording an event and recording moments of memory, and with the advent of social media to share their lives. Figure \ref{fig:top20} shows the award-winning photos of 2020\footnote{\url{https://www.photoawards.com/winner/?compName=IPA+2020}} from different realms of life. The era of Instagram and Flickr has seen an unprecedented overload of digital photography. According to the latest statistics shared by Omnicore Agency, $50B+$ photos have been uploaded on the platform and $995$ photos are uploaded every other second\footnote{\url{https://www.omnicoreagency.com/instagram-statistics/}}, $350M$ photos are uploaded every day on Facebook\footnote{\url{ https://www.omnicoreagency.com/facebook-statistics/}}, and a decade is needed to view all photos on Snapchat\footnote{\url{https://www.omnicoreagency.com/snapchat-statistics/}}. Such an upsurge in the proliferation of data has given rise to various computer vision applications.

Image classification and recognition have revolutionized the data visualization happening digitally. The ability to automatically identify objects has led to the development of various applications like image tagging, organization, categorization, behavioral analysis, recommendation systems, and so on. A dataset of images with various photographic styles is useful to carry out style-based image retrieval that can narrow-down image search, image recommendation based on preferences, generation of royalty-free synthetic images to avoid copyright restrictions, and many more applications.

In this paper, we present \emph{Photozilla}, a large-scale dataset of $990k$ million images comprising $10$ different photography styles. To demonstrate the usefulness of this dataset, we propose $3$ photo-style classification models that outperform state-of-the-art classifiers and achieve over $\sim96$\% accuracy for predicting the correct photography style. 

While photography is a rapidly evolving subject with many new styles being introduced over the years, our models must be able to quickly adapt to identify these new styles. The conventional deep learning classification models require large data sets to achieve a high level of accuracy. To overcome this issue and to enable our models to be able to quickly adapt to a new photography style even with a few samples, we propose to use few-shot learning. We implemented a Siamese network where we use our classification models as base architecture to quickly adapt to new styles with only $25$ training samples. 

Henceforth, the following are the contributions from this work:
\begin{enumerate}
    \item We present a large-scale photography dataset comprising over $990k$ images under $10$ different photography styles. 
    
    \item By using three state-of-the-art classification models, we achieve an accuracy of $\sim96$\% on classifying to $10$ these styles.
    
    \item We further extend our dataset with $10$ additional photography styles but with a small number of samples ($25$ per class). By using the pre-trained $10-$class classification model and a Siamese architecture, we show that our model can learn to classify new photography styles even with even a low number of samples. We achieve an accuracy of $\sim68$\% for the same.
\end{enumerate}

The remainder of the paper is organised as follows. Section \ref{sec:related} presents a review of the literature. Section \ref{sec:dataset} discusses the procedure of dataset accumulation. Section \ref{sec:methodology} describes the methodology in detail, while experimental results are shown in Section \ref{sec:experiments}. The future work is discussed in Section \ref{sec:disc} and concluding remarks are given in Section \ref{sec:conc}.

\section{Related Works}
\label{sec:related}
With the breakthroughs in Deep Learning technology, photography has turned out to be an area of constant interest in the research community. From analyzing images for interpretation to synthetic image generation, an immense amount of applications has been developed. With these applications, many researchers have published various kinds of datasets in photography for specific uses and developed many state-of-the-art computer vision techniques. Some of the notable works are mentioned below.

\textbf{Photography Datasets.}~\cite{murray2012ava} published a large-scale dataset, AVA, to perform Aesthetic Visual Analysis (AVA) with $\sim250k+$ images containing aesthetic scores of all data points, semantic labels for over $60$ categories, and labels for $14$ photographic styles based on light, color, and composition. \cite{karayev2013recognizing} created two datasets: (a) $80k$ photographs from Flickr \cite{flickr} with corresponding style annotation, and (b) $85k$ paintings with $25$ styles/genre annotation in order to predict image style and aesthetic standard. The photographic styles were categorized based on photographic technique, composition style, mood, genre, and type of scenes. The work aimed to show the significance of an image style in the age of digital photo-overload. 

Some researchers have worked in specific genres, like \cite{khan2014painting}, who proposed a dataset of digital paintings with more than $4.2k$ paintings from $91$ painters having two annotation labels: artist name and style. \cite{wilber2017bam} focused on contemporary artwork and featured an artistic imagery dataset, BAM, collected from the Behance website. The collected dataset was annotated with labels for content, emotions, and artistic media. Other notable datasets that primarily emphasized artworks are \cite{wikiart, zhang2019imet}.

Food imagery has gained substantial attention because of its multi-purpose applications including health, food selection and recommendation, culture, and so on \cite{min2019survey}. With this motivation, \cite{sheng2018gourmet} presented a Gourmet Photography Dataset (GPD) having $12k$ photographs to assess the aesthetics in food images. \cite{sahoo2019foodai} introduced a large-scale food image dataset with $152$ categories consisting of generic types of food and $756$ visual food items constituting a total of $\sim$400k+ images. 

Some other genre-specific datasets include scenery database \cite{patterson2012sun}, aerial images dataset \cite{xia2018dota, chen2020valid}, and street-level images dataset \cite{neuhold2017mapillary}.

\textbf{Image Classification.} A considerable increase in the popularity of image classification tasks can be noticed since the introduction of the revolutionary ImageNet dataset \cite{deng2009imagenet}. Developed with the aim of visual object recognition, ImageNet at present contains $14M+$ annotated images of $20k+$ specific object categories with bounding boxes in $1M+$ images. Another large-scale visual dataset is the COCO ( Common Object in Context) dataset \cite{lin2014microsoft}. It comprises instance segmentation of 80 common objects. The dataset is containing $328K$ images of $2.5M$ labeled instances. 

The aforementioned datasets engendered state-of-the-art neural network architectures for image classification. A $152-$layered network, ResNet (Residual Network) \cite{he2016deep} won the $2015$ ImageNet challenge with the introduction of a novel concept of skip connection. These connections were designed to solve the problem of vanishing gradients caused by skipping more than one layer in the network and hence, use the output of one layer as the input for others. This enables the easier flow of the gradient from layer to layer. One of the popular variants of ResNet is DenseNet \cite{huang2017densely} where the concept of extra connections was introduced to resolve the same problem. All the layers with identical feature-maps in the network are directly interconnected with each other so that maximum information flow will take place. 

The classification models we use, \cite{zagoruyko2016wide, xie2017aggregated, tan2019efficientnet} are discussed in Section \ref{sec:classfication}.

\textbf{Classification Similarity Learning.} Similarity-based classification takes pairwise input images and predicts a similarity score, instead of classifying an image directly as any target class. This score can be in a form of binary value i.e. $0$ or $1$, or any real number. To train a network to learn similarity, Siamese Neural Networks \cite{koch2015siamese} were introduced. More details can be found in \ref{subsubsec:siamese}. Siamese networks can be further used to perform low-shot learning in which a limited number of samples are used to train the model. Some examples are zero-shot \cite{larochelle2008zero}, one-shot \cite{miller2000learning}, and few-shot learning (FSL) in which zero, one, and a few training samples are provided respectively. More details can be found in \ref{subsec:fewshot}.

\section{Dataset Collection}
\label{sec:dataset}
\begin{figure}[h]
    \centering
    \includegraphics[width=\linewidth]{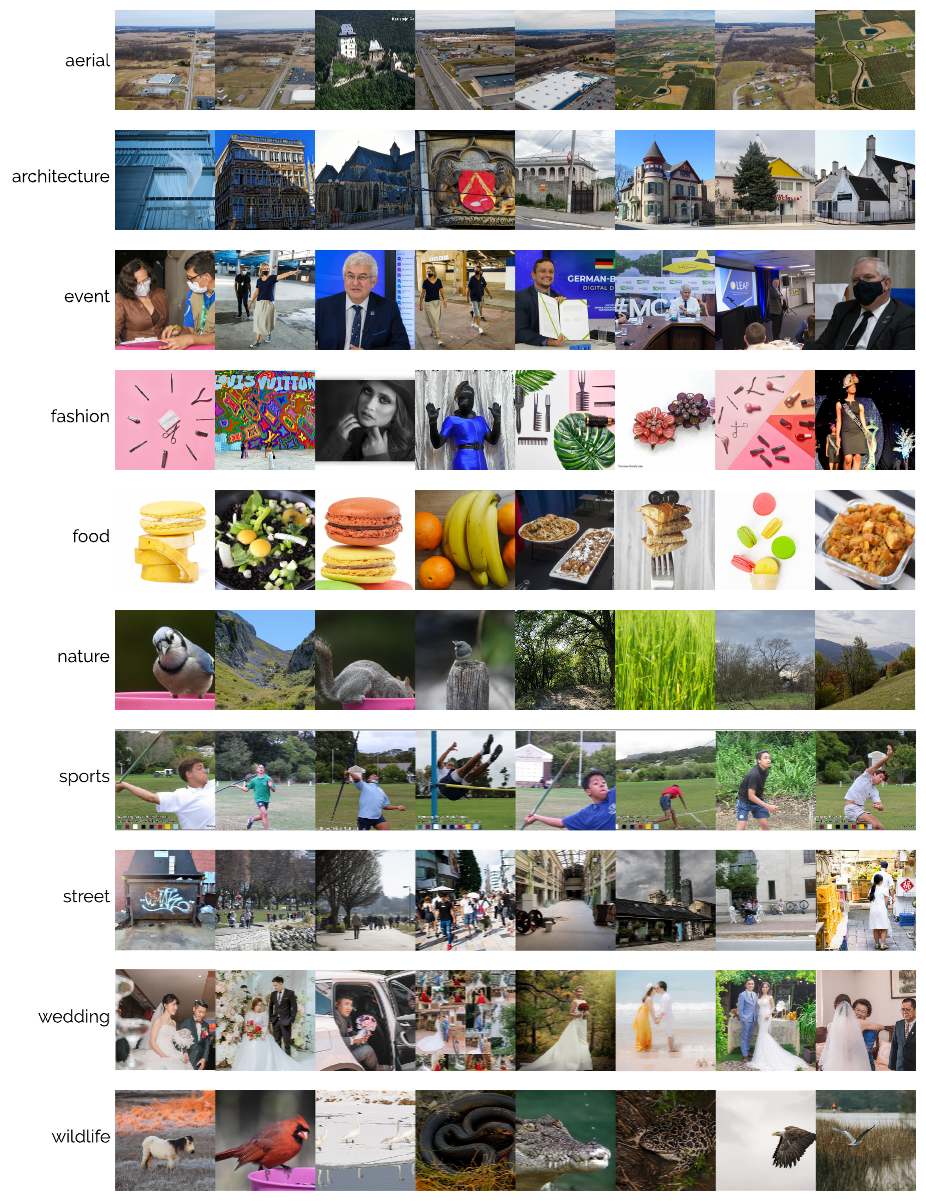}
    \caption{Samples images from $10$ classes of the training dataset for classification models.}
    \label{fig:10train}
\end{figure}

For our model evaluation, we first collected a large-scale dataset of Flickr\footnote{\url{https://www.flickr.com/}} images using Flickr API to automatically extract the images belonging to the $10$ photography styles. These images of specific styles were collected using tags of the same. For example, to crawl images in travel photography, \textit{`travel'} was used as the specific tag. Each class of $10$ has approximately $\sim100k$ number of images. Figure \ref{fig:10train} shows samples of each  photography style.  

\begin{figure}[h]
    \centering
    \includegraphics[width=\linewidth]{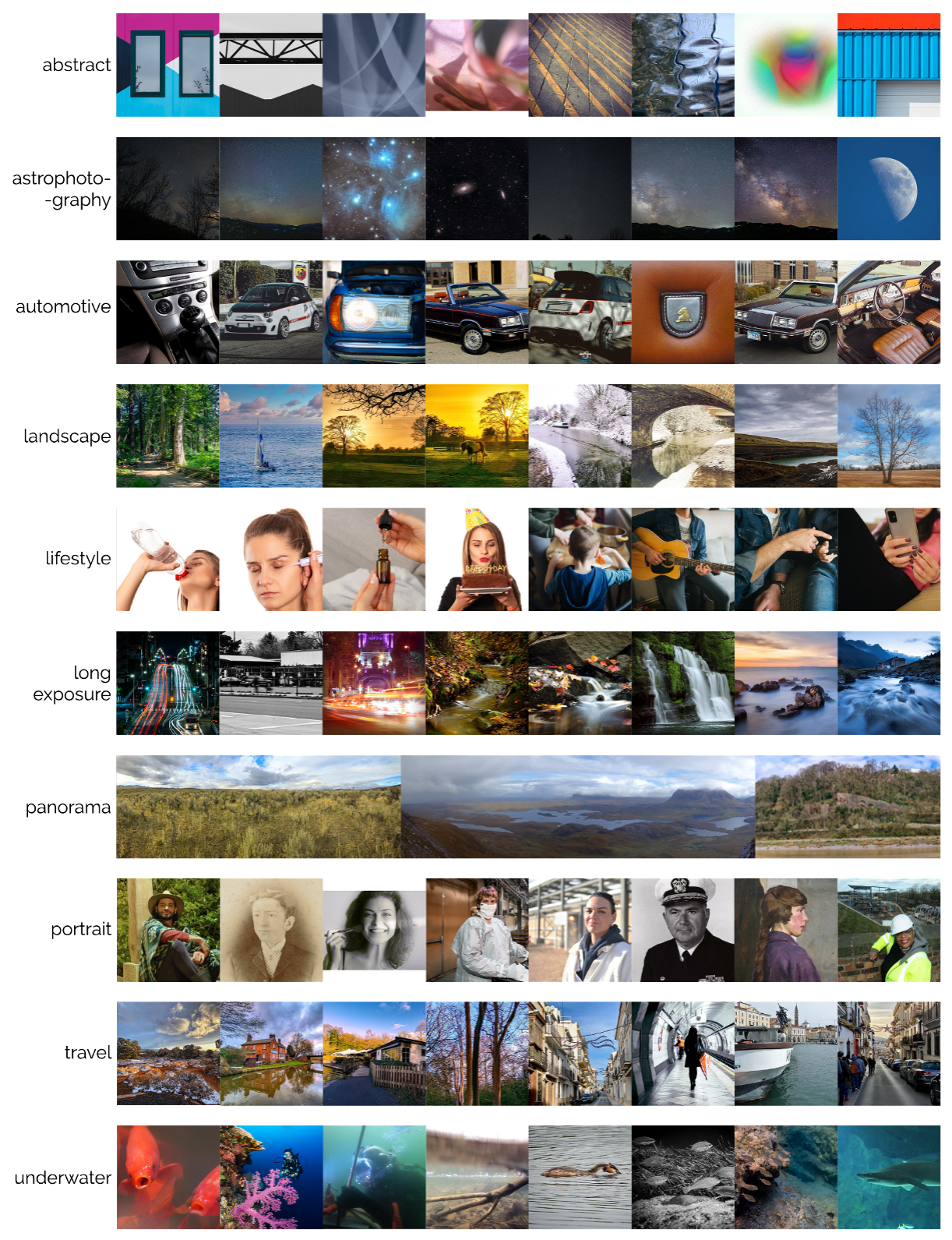}
    \caption{Samples images from Siamese network' testing dataset of 10 classes.}
    \label{fig:10test}
\end{figure}

We extended our dataset with $10$ more classes to our dataset, as shown in figure \ref{fig:10test}, but with only a limited number of data samples i.e. $25$ per class. This extended dataset was then used for our evaluations of few-shot learning of unseen photography classes.  

\section{Methodology}
\label{sec:methodology}

\begin{figure*}[h]
    \centering
    \includegraphics[width=\linewidth]{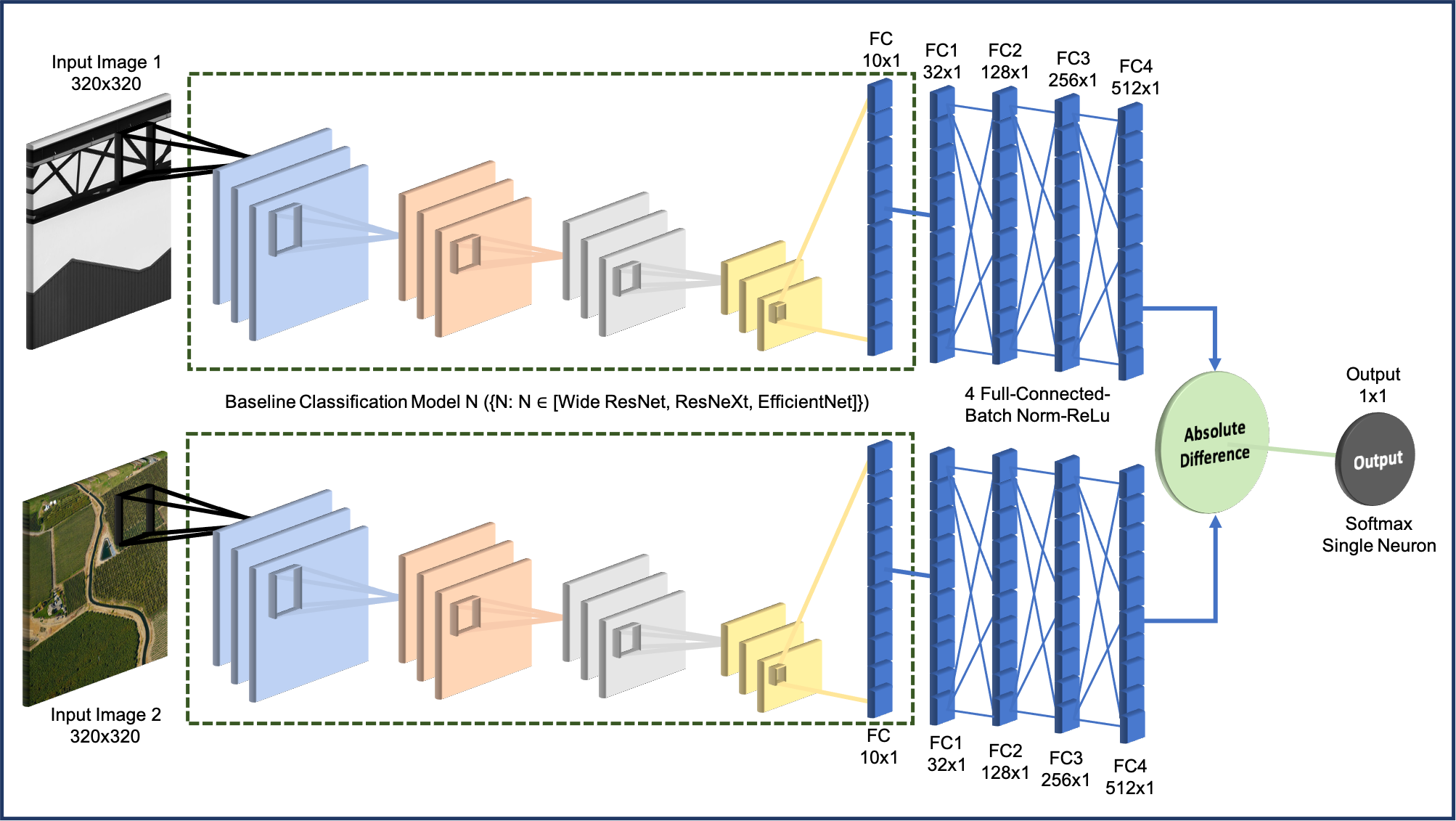}
    \caption{Proposed siamese-based neural network architecture.}
    \label{fig:nn}
\end{figure*}

\subsection{Classification Models}
\label{sec:classfication}
In our proposed work, three state-of-the-art classification models are used i.e. Wide ResNet \cite{zagoruyko2016wide}, ResNext \cite{xie2017aggregated}, and EfficientNet \cite{tan2019efficientnet} to train the baseline visual embeddings with the 10 classes with a larger number of datapoints. For all classification models, we have used identical hyper-parameters as listed in Table \ref{tab:hyperparam}. The classification accuracies on the curated dataset are shown in the Table \ref{tab:class3}.

\begin{table}[h]
\centering
\caption{Hyperparameters used for classification models.}
\begin{tabular}{@{}cc@{}}
\toprule
Hyparameter   & Value              \\ \midrule
No. of Epochs & 1                  \\
Optimizer     & Stochastic Gradient Descent                \\
Learning Rate & 0.01               \\
Momentum      & 0.9                \\
Batch Size    & 64                 \\
Loss Function & Cross Entropy Loss \\ \bottomrule
\end{tabular}
\label{tab:hyperparam}
\end{table}

\subsubsection{\textbf{Wide ResNet}}

In the earlier work on Deep Neural Networks, performance would improve when more layers were stacked up. However, performance gains appear to saturate beyond a certain point. To mitigate this issue of performance saturation, Deep Residual Networks (ResNet) \cite{he2016deep} were introduced. These deeper networks achieved superior performance while being able to scale up to thousands of layers. The key intuition of such architectures is the usage of residual blocks, which is primarily represented by the following equation:

\begin{equation}
    \label{eq:residual}
    x_{l+1} = x_{l} + \mathcal{F}(x_l,\mathcal{W}_l)
\end{equation}

Here, $x_l$, $\mathcal{W}_l$ and $x_{l+1}$ are the input, weights and the output of the $l$\textsuperscript{th} layer, respectively. $\mathcal{F}$ is the residual function governed by the architecture of the residual block.

Although ResNet architectures were able to achieve superior performances in comparison to the shallower models, these performance gains came at the cost of increased training and inference times. To overcome these issues, Wide Residual Networks (WRNs) \cite{zagoruyko2016wide} were introduced. WRNs have a nearly similar model architecture as ResNet except for the increased number of feature maps and shallower architecture. They have explored the benefit of increasing the width of the network by a hyper-parameter $k$ instead of a deeper architecture.

\subsubsection{\textbf{ResNeXt}}
Instead of having higher depth and width, ResNext \cite{xie2017aggregated} uses the concept of having higher cardinality, a new dimension introduced to improve the performance of networks with less complex architectures. The cardinality, $C$ of a model can be defined as the number of branches in a residual block to control more complex transformations. Mathematically, these $C$ transformations are formulated as follows. 

\begin{equation}
    \label{eq:resnext}
    \mathcal{F}(x_l,W_l) = \sum _{i=1}^{C} \mathcal{T}_i(x_l,{W_l}^i)
\end{equation}

Here, $T_i$ is the transformation function for the branch $i$ of the residual block. The aggregated transformation is the sum of all $C$ branches. This aggregrated transformation is then used as the residual function similar to equation \ref{eq:residual}.

\subsubsection{\textbf{EfficientNet}}
EfficientNet \cite{tan2019efficientnet} uses a novel scaling method that considers three dimensions of a neural network: depth, width, and resolution. EfficientNet performs compound scaling that combines the scaling of all three dimensions to optimize for accuracy while meeting the memory and computational constraints. To reduce the design space, all layers must be scaled uniformly with a constant ratio. Let $d$, $w$ and $r$ be these constant ratios for depth, width and resolution, respectively. Let $\mathcal{N}(d,w,r)$ be the resulting neural network with these ratios. EfficientNet proposes the following optimization function to find optimal $d$, $w$ and $r$. 

\begin{equation}
\begin{split}
  \max_{d,w,r} Accuracy(\mathcal{N}(d,w,r)) \\
  Memory(\mathcal{N}) \leq target\_memory \\
  FLOPS(\mathcal{N}) \leq target\_flops
\end{split}
\end{equation}

\subsection{Few-Shot Learning}
\label{subsec:fewshot}
Existing deep neural network architectures generally contain billions of parameters. Thus, training such DNN architectures requires a large training set which is often challenging due to the data collection and annotation of the ground truth. Few-Shot learning, as the name suggests, is a class of techniques that allows a DNN to learn with a smaller number of training samples. In our canonical case, photography is a rapidly evolving industry. Over time, various styles of photography have evolved and it is infeasible to modify our base classification model to adapt to a newer photography style. Instead, we use a network architecture called Siamese network to quickly adapt our base classification model to an unseen class of photography even with a low number of annotated training samples.

\subsubsection{\textbf{Siamese Network}}
\label{subsubsec:siamese}
Siamese networks \cite{koch2015siamese} are a class of deep neural networks that contains two identical sub-networks. Here, identical means that the sub-networks have the same architecture, parameters, and weights. Traditionally, classification networks learn to classify a training sample into multiple classes. Siamese networks, however, learn a deep similarity function that takes two different inputs and computes whether the inputs belong to the same class or different classes. The two identical sub-networks each take one input and compute a deep feature embedding for that input. This deep feature embedding is then used to compute a similarity score to determine whether the two inputs are from the same class or two different classes.

As mentioned in the previous section, we have trained three state-of-the-art classification models for predicting the style of an image as being one of the $10$ photography styles. We use the output of the last fully connected layer of these base classification models and then stack $4$ more fully connected layers to compute the $512-$dimensional deep feature embedding. Finally, we compute the similarity score by computing the absolute difference between the two deep feature embeddings from siamese. Let $I_1$ and $I_2$ be two images for which we measure the similarity score. Let $\mathcal{N}(I)$ be the 512-dim visual feature embedding output for image $I$. Then, we compute the similarity score $P(I_1,I_2)$ as follows.

\begin{equation}
\begin{split}
    P(I_1,I_2) = Softmax(W_{out}\cdot|\mathcal{N}(I_1)-\mathcal{N}(I_2)|+B_{out}) \\ 
    Where \;\; W_{out} \in \mathcal{R}^{512 \times 1} \;\; and \;\; B_{out} \in \mathcal{R}^{1 \times 1}
\end{split}
\end{equation}

Furthermore, we use the cross-entropy loss in our Siamese network and a learning rate of $0.05$ with SGD \cite{ruder2016overview} as our optimizer to train for $30$ epochs with each class having only $25$ training, validation, and test samples each. 

Figure \ref{fig:nn} represents the complete model architecture of the proposed work.


\section{Experimental Results}
\label{sec:experiments}

\subsection{\textbf{Classification Accuracy}}

As explained in Section \ref{sec:classfication}, we used $10$ photography classes to evaluate the classification accuracy of our dataset on $3$ state-of-the-art models (see \ref{tab:class3}). $70\%$ of the dataset was used for training and $30\%$ as the test dataset. All three models achieve over $96\%$ accuracy on the test dataset. ResNeXt marginally outperforms the other two with an accuracy of $96.35\%$.

\begin{table}[h]
\centering
\caption{Classification models' accuracies.}
\begin{tabular}{@{}cc@{}}
\toprule
Classification Model & Accuracy (\%) \\ \midrule
Wide ResNet          & 96.23         \\
ResNeXt              & 96.35         \\
Efficient Net        & 95.71         \\ \bottomrule
\end{tabular}
\label{tab:class3}
\end{table}

\subsection{\textbf{10-Way Few-Shot Evaluation Metric}}

We further extracted $75$ images each for $10$ additional photography classes. These additional classes were used for evaluating the performance of our proposed Siamese network for few-shot learning of new photography classes. Out of these $75$ images, $25$ each was used for training, validation, and testing respectively.

For the evaluation of the test dataset, we used the $10-$Way few-shot evaluation metric. In this evaluation task, we take one image $Q$ belonging to class $c$ as the query image, and randomly pick $10$ more images from each class. Let $I_j$ be the randomly picked image for $j$\textsuperscript{th} class. For the pair of images $Q$ and $I_j$, the Siamese network predicts a similarity probability $P(Q,I_j)$, which defines the similarity between two images. We run the Siamese network for image $Q$ and $I_j \forall j \in [1,10]$ and select the image with the highest $P(Q,I_j)$. A prediction is said to be correct if the following criterion is met. 

\begin{equation}
    c = argmax_{j \in [1,10]}(P(Q,I_j))
\end{equation}

Table \ref{tab:siamese} depicts the $10-$Way few-shot evaluation accuracies for siamese with different base networks. In summary, the Siamese network with ResNext performs better than the other two variants ($68.34\%$). EfficientNet is currently the state-of-the-art network for image classification. Surprisingly, EfficientNet only gives an accuracy of $60.84\%$. The other two variants give an ~$8-12\%$ higher accuracy.

\begin{table}[h]
\centering
\caption{Siamese network' accuracies with different classification models.}
\begin{tabular}{@{}cc@{}}
\toprule
Model (with Siamese) & Accuracy (\%) \\ \midrule
Wide ResNet          &  64.17   \\
ResNext              &  68.34   \\
EfficientNet         &  56.25   \\ \bottomrule
\end{tabular}
\label{tab:siamese}
\end{table}

\subsection{Qualitative Analysis of Visual Embedding for clustering}

To further analyze the capability of our Siamese network to classify unseen photography styles with only a few samples, we use t-Distributed Stochastic Neighbor Embedding (t-SNE) \cite{van2008visualizing}. t-SNE is a non-linear and unsupervised dimensionality reduction approach to visualize high-dimensional data. Intuitively, t-SNE allows one to visualize how the data is arranged in the higher dimensional space. 

\begin{figure*}[]
    \centering
        \begin{subfigure}[b]{0.488\textwidth}
            \centering
            \includegraphics[width=1\textwidth]{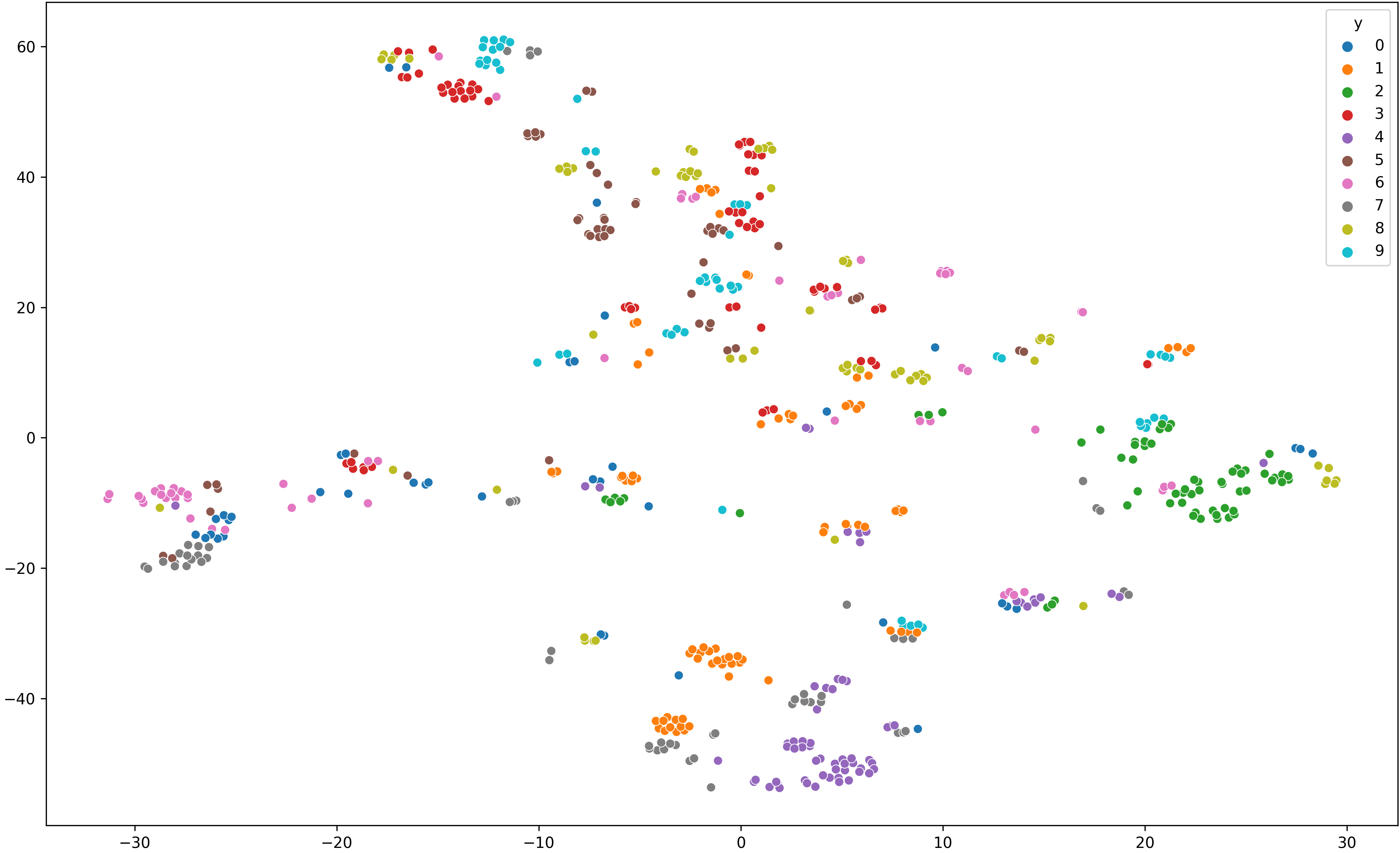}
            \caption[(a)]%
            {{\small}}    
        \end{subfigure}
        \hfill
        \begin{subfigure}[b]{0.488\textwidth}  
            \centering 
            \includegraphics[width=1\textwidth]{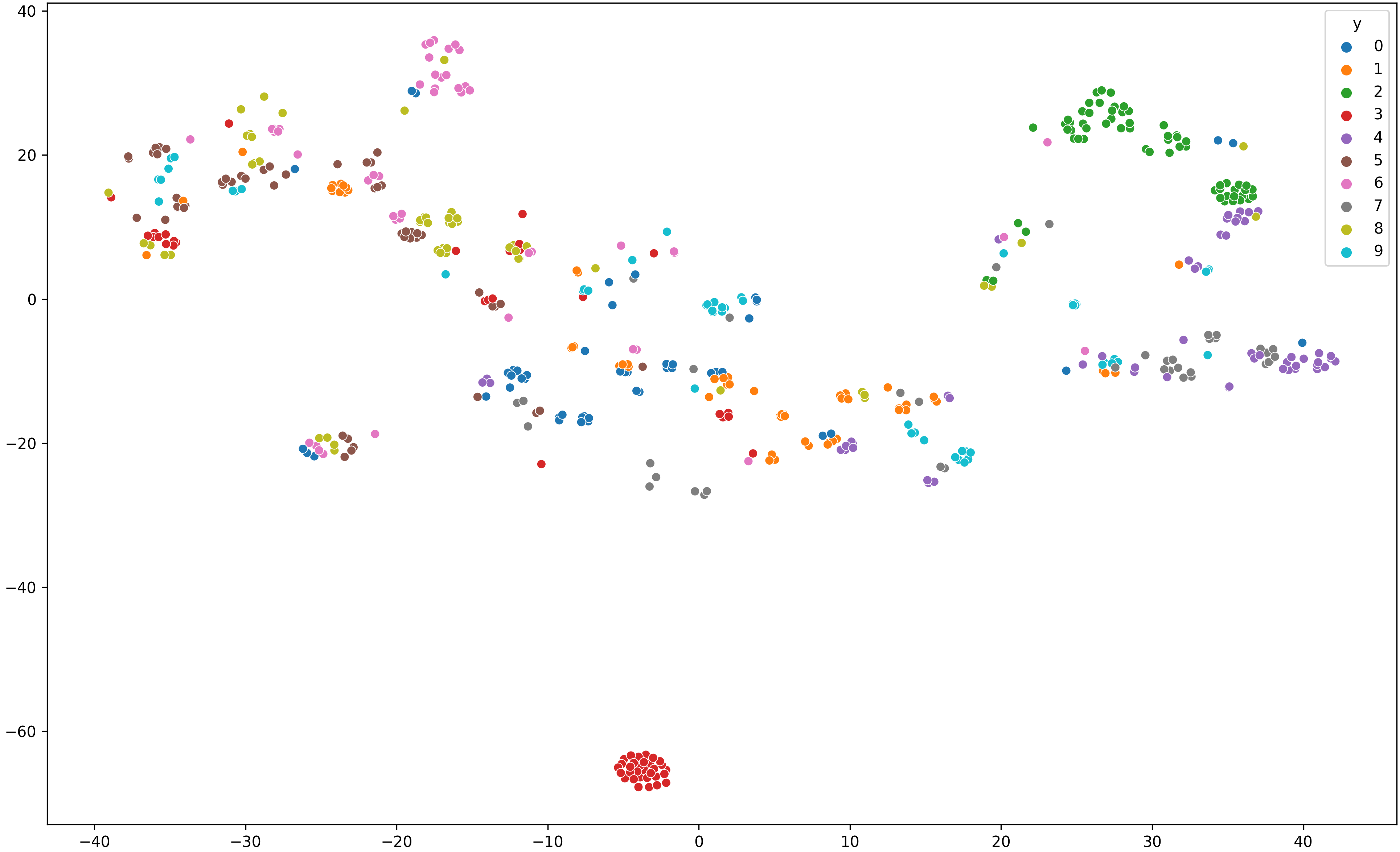}
            \caption[(b)]%
            {{\small}}    
        \end{subfigure}
        \begin{subfigure}[b]{0.488\textwidth}   
            \centering 
            \includegraphics[width=\textwidth]{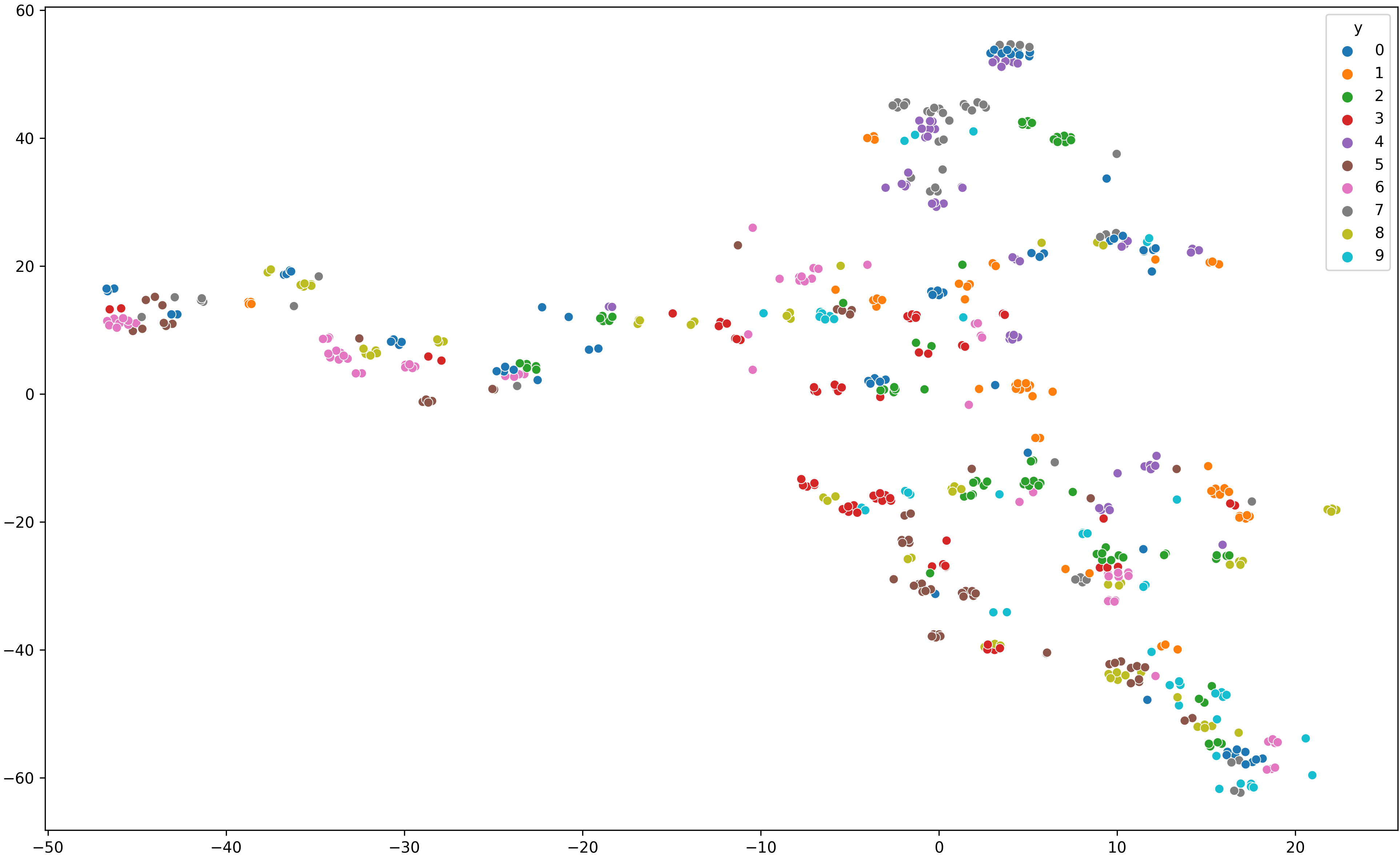}
            \caption[(c)]%
            {{\small}}    
        \end{subfigure}
    \caption{Clusters Visualization using t-SNE for (a) Wide ResNet, (b) ResNeXt, and (c) EfficientNet. Here, 'y' represents the different photography styles.}
   \label{fig:clusters}
\end{figure*}

We use the 512-dim feature embedding output as an input to generate the t-SNE transform. Figure \ref{fig:clusters} shows the t-SNE plots to visualize the 512-dim visual embedding arranged in a 2-dim space. 

Based on this, we can observe that the Siamese variant of ResNeXt and Wide ResNet can generate better clusters for images of the same photography style. However, we do not observe any distinct clusters for EfficientNet. This is also proven by the performance comparison of Siamese networks where ResNeXt and Wide ResNet perform comparatively better than EfficientNet's Siamese network.   

\section{Discussion and Future Work}
\label{sec:disc}

In this section, we discuss some potential applications of this work and highlight future directions for research.

\subsection{Photozilla as a Service}
Our proposed model can automatically identify $10$ different classes and even adapt to new and unseen photography styles with merely $25$ training samples. We envision our model as an API service to be used by digital photography platforms. We believe that these platforms will greatly benefit from such a service as it allows the automatic curating of their collection with relevant tags, enables photograph recommendation systems, etc. 

\subsection{Siamese Visual Embedding for Photography Recommendation}

In our proposed Siamese network, we use a $512-$dim visual feature embedding to quantify the similarity between two images. We could instead use this embedding for recommending images based on user's preferences. Such a system could use user-specific features, e.g. user's biographies, previous views, and likes, to generate another feature embedding. This feature embedding can be compared with the image embedding to recommend users with photographs of their preference. Mathematically, let us define $\mathcal{N}_V(I)$ to be the network that takes image $I \in \mathcal{I}$ as an input to generate visual feature embedding. Let $\mathcal{N}_T(U)$ be the network that takes the user's features $U$ as an input to generate user-specific feature embedding. The system could recommend an image $I$ to the user if, $dist(\mathcal{N}_V(I),\mathcal{N}_T(U)) \le \mathcal{T}$. Here, $\mathcal{T}$ is a pre-determined similarity threshold.

\subsection{Royalty-Free Images}

Currently, with Photozilla, we have collected a large-scale dataset of images under Creative Commons license and copyright-free images. Moving forward, we intend to utilize this dataset to generate synthetic images using state-of-the-art generative models. Such royalty and copyright-free synthetic images could be beneficial for the academic communities to carry out various researches.

\subsection{Next-POI Recommendation}
We intend to collect more information and fuse different types of features, such as captions, locations, and hashtags ~\cite{heng2020urban,liu2020epic30m}, and perform multi-modal analysis and sequence prediction tasks for various applications of the dataset, such as next-POI (Point-Of-Interest) recommendations~\cite{liu2020strategic}.

\subsection{Image Quality and Aesthetic Assessment}
Finally, we aspire to collaborate with experts from the design community to propose models and conduct experiments to address the non-trivial issue of aesthetic and quality assessment.

\section{Conclusion}
\label{sec:conc}
This work presents a large-scale dataset termed \textit{`Photozilla'} comprising over $990k$ images belonging to $10$ photography styles. We used this dataset as a canonical example to train $3$ different classification model architectures to automatically identify the photography style. These models achieve superior performance of over $96\%$ accuracy on our testing dataset. Digital photography is a rapidly evolving field, which requires that our models can adapt quickly to identify new photography styles. To facilitate this, we propose a novel Siamese network that learns from our base classification networks. The proposed Siamese network achieves an accuracy of over $68\%$ on identifying $10$ new photography styles with merely $25$ training samples.






\end{document}